\documentclass[journal]{IEEEtran}
\ifCLASSINFOpdf
\else
\fi
\usepackage{times}
\usepackage{soul}
\usepackage{url}
\usepackage[hidelinks]{hyperref}
\usepackage[utf8]{inputenc}
\usepackage[small]{caption}
\usepackage{graphicx}
\usepackage{amsmath}
\usepackage{amsthm}
\usepackage{amsfonts}
\usepackage{array, booktabs}
\usepackage{algorithm}
\usepackage{algorithmic}
\usepackage[switch]{lineno}
\usepackage{multirow}
\usepackage{subcaption}
\usepackage{makecell}
\usepackage{float}

\let\oldFootnote\footnote
\newcommand\nextToken\relax

\renewcommand\footnote[1]{%
    \oldFootnote{#1}\futurelet\nextToken\isFootnote}

\newcommand\isFootnote{%
    \ifx\footnote\nextToken\textsuperscript{,}\fi}

\usepackage{color}

\begin{document}
%
\title{Bridging Large Language Models and Optimization: A Unified Framework for Text-attributed Combinatorial Optimization}
%
%
%

\author{Xia Jiang,
        Yaoxin Wu,
        Yuan Wang,
        and
        Yingqian Zhang
\thanks{Xia Jiang, Yaoxin Wu, and Yingqian Zhang are with the Department of Industrial Engineering and Innovation Sciences, Eindhoven University of Technology, 5600 MB Eindhoven, The Netherlands (email: summer142857.jiang@gmail.com; wyxacc@hotmail.com; yqzhang@tue.nl). Yuan Wang is with Shenzhen Research Institute of Big Data, Shenzhen, China (email: wangyuan@cuhk.edu.cn)}
}

%
%

\markboth{Bridging Large Language Models and Optimization: A Unified Framework for Text-attributed Combinatorial Optimization}%
{Shell \MakeLowercase{\textit{et al.}}: Bare Demo of IEEEtran.cls for IEEE Journals}
%



\maketitle


\begin{abstract}
To advance capabilities of large language models (LLMs) in solving combinatorial optimization problems (COPs), this paper presents the Language-based Neural COP Solver (LNCS), a novel framework that is unified for the end-to-end resolution of diverse text-attributed COPs. LNCS leverages LLMs to encode problem instances into a unified semantic space, and integrates their embeddings with a Transformer-based solution generator to produce high-quality solutions. By training the solution generator with conflict-free multi-task reinforcement learning, LNCS effectively enhances LLM performance in tackling COPs of varying types and sizes, achieving state-of-the-art results across diverse problems. Extensive experiments validate the effectiveness and generalizability of the LNCS, highlighting its potential as a unified and practical framework for real-world COP applications.
\end{abstract}


%
\IEEEpeerreviewmaketitle

\section{Introduction}

Large language models (LLMs) have revolutionized artificial intelligence, demonstrating remarkable capabilities across various domains such as language understanding \cite{laskar2023systematic, qian2024tell}, code generation \cite{chen2021evaluating, zhang2023repocoder} and mathematical reasoning \cite{chen2024masked, zhao2024docmath}. 
Recently, there has been a growing trend in investigating the application of LLMs in automatically solving combinatorial optimization problems (COPs).

COPs represent a fundamental class of mathematical problems that focuses on finding optimal solutions from finite sets of objects. Typical examples of COPs include the traveling salesman problem (TSP), the vehicle routing problem (VRP), the knapsack problem (KP), etc., which are challenging to solve due to the NP-hardness. COPs have been approached using exact methods or (meta-)heuristics in operations research \cite{Blum2003}, while recent neural combinatorial optimization (NCO) methods have emerged as promising alternatives \cite{bengio2021machine}. However, these methods require domain expertise and specialized designs for each specific COP, hindering their applications to diverse COPs. In contrast, recent research has turned to LLMs to tackle COPs through natural language interactions (i.e., prompting), with the aim of further enhancing the automation of the problem solving process \cite{yang2023large, masoud2024exploring, iklassov2024self}.

However, achieving high-quality solutions solely by prompting LLMs remains challenging, even for middle-sized COPs (e.g., those with above 50 nodes) \cite{yang2023large}. Studies have indicated that LLMs often struggle with processing problems involving lengthy contexts, as they tend to lose coherence and accuracy in extended language descriptions \cite{xu2024can}. Furthermore, LLMs often struggle to effectively comprehend and represent the intricate relationships between problem elements, such as graph structures in COPs. The inherently verbose and potentially ambiguous nature of COP descriptions in natural language raises considerable concerns about the ability of LLMs to independently generate satisfactory solutions.

Despite these challenges, LLMs have gained considerable advantages in developing sophisticated linguistic knowledge and semantic representation, through their pretraining \cite{yu2023empower}.
Their ability to represent natural language opens opportunities for integration with additional neural networks to address downstream text-based tasks, such as node classification and link prediction on text-attributed graphs \cite{zolnai2024stage, liu2024one}. Specifically, LLMs can also encode descriptions of COPs, aligning them in a unified semantic space. A much lighter-weight network (compared to LLMs) can then leverage the aligned embeddings to generate solutions across problems.

This paper presents the \textbf{L}anguage-based \textbf{N}eural \textbf{C}OP \textbf{S}olver (LNCS), a novel framework that integrates LLMs with a Transformer-based neural network to process text-attributed COPs and generate solutions. 
Concretely, it utilizes LLMs to encode text-attributed instances (TAIs) of different COPs into a shared semantic space. The semantic embeddings of TAIs are further processed by a Transformer network, which learns solution construction rules to generate solutions. By training the Transformer with reinforcement learning (RL) while freezing LLM parameters, LNCS borrows the LLM's capability for generic representation of problem descriptions, and gains satisfactory performance to solve various COPs in a unified manner.

Our main contributions are threefold: 1) we propose the LNCS to synthesize LLMs' semantic representations with a Transformer network for solution generation, bridging LLMs and text-attributed COPs for end-to-end problem resolution; 2) we incorporate a conflict-free multi-task RL algorithm to facilitate the training for diverse COPs with varying scales of gradients through a unified model; 3) we evaluate LNCS on different text-attributed COPs, showcasing that LNCS outperforms typical prompting and optimization approaches. Moreover, we extensively validate the effectiveness and generalizability of LNCS, which can be favorably fine-tuned for COPs of varying sizes and types.


To the best of our knowledge, this work represents the first successful application of LLMs to build a unified (Transformer) model for solving general text-attributed COPs.
It creates a vital connection between the linguistic capabilities of LLMs and their enhanced performance in COPs tasks.

\section{Related work}
\subsection{LLM for Optimization}
Research on the use of LLMs for optimization tasks has progressed along two main paradigms: approaches that use \textit{LLMs as programmers} to generate solution functions and those that employ \textit{LLMs as optimizers} to directly produce solutions.

\noindent\textbf{LLMs as Programmers.} 
Recent advances showcase that LLMs are capable of writing programs to implement heuristic algorithms to solve COPs \cite{romera2024mathematical, iklassov2024self}. Starting from an initial code template, LLMs can iteratively refine heuristics through an evolutionary process \cite{fei2024eoh, ye2024reevo}. However, the development of effective algorithms through this evolutionary process often necessitates substantial domain knowledge and token consumption for each specific COP. A more pragmatic approach enables LLMs to interface with established optimization solvers, such as Gurobi and OR-tools \cite{xiao2024chainofexperts, zhang2024solving}. These methods generally focus on formulating a COP instance using mathematical programming models \cite{wasserkrug2024large}, but they struggle to guarantee the correctness of the generated formulations, and thus are not applicable to real world scenarios.

\noindent\textbf{LLMs as Optimizers.} LLMs can also function as black-box optimizers, which directly generate solutions \cite{abgaryan2024llms} or iteratively refine an initial solution \cite{yang2023large,liu2023large11}. Prompting techniques are pivotal in these approaches to solving COPs from language descriptions. However, current methods still present a substantial research gap in achieving high-quality solutions \cite{iklassov2024self}. Research by \cite{zhang2024llmgraphreasoninggeneralize} suggests that LLMs tend to memorize limited patterns in training data rather than develop generalizable reasoning skills, which impedes their effectiveness as direct optimizers.


In this paper, the LNCS brings a novel paradigm to integrate LLMs with a Transformer network. Compared to the above paradigms, the LNCS enables LLMs to optimize diverse COPs through a single solution generator (i.e., the Transformer network), and enhance their capacity to solve COPs.

\subsection{Neural Combinatorial Optimization}
The constructive NCO methods aim to learn policies to construct solutions in an autoregressive way. The early attempts are based on pointer networks \cite{vinyals2015pointer, bello2016neural}, a class of recurrent neural networks (RNNs) that process the input and generate the solution in a sequence-to-sequence fashion. Inspired by the Transformer \cite{vaswani2017attention}, the attention model (AM) is presented to solve VRPs, respectively, showing the advantage over conventional heuristics. Afterward, a series of strategies are proposed to enhance Transformer-based NCO models by leveraging the symmetricity of COPs \cite{kwon2020pomo,kim2022sym,fang2024invit} and efficient active search \cite{hottung2021efficient,qiu2022dimes,choo2022simulation}. In addition, the improvement NCO methods enhance stochastic search algorithms, such as local search \cite{9393606,hudson2021graph}, neighborhood search \cite{ijcai2022p662,li2021learning} and evolutionary algorithms \cite{ye2024deepaco}, by employing neural networks to iteratively improve an initial solution.


In contrast to NCO methods, the LNCS framework aims to address general text-attributed COPs described by natural language. It provides a single model for diverse COPs by leveraging LLMs' unified semantic representations of texts.


\section{Preliminaries}

\subsection{Combinatorial Optimization Problems}
Solving a COP involves searching for an object within a finite (or countably infinite) discrete set, where
the object can be an integer, a subset, or a permutation \cite{Blum2003}. 
Most COPs could be represented on graphs in which objects are denoted by nodes and edges. More formally, a COP $P$ is formulated as follows:
\begin{equation}
    \min\limits_{\boldsymbol{x}} f(\boldsymbol{x}, P) \quad s.t. \scalebox{0.5}{\quad} c_j(\boldsymbol{x}, P) \leq 0, j=0,1,...,J
\end{equation}
where $\boldsymbol{x}=\{x_1,...,x_n\}$ is a set of discrete decision variables, defined by $\{x_i\in D_i\}_{i=1}^n$; $f(\boldsymbol{x}, P)$ denotes an objective function to be minimized and $\{c_j(\boldsymbol{x}, P)\}_{j=1}^J$ denotes a set of problem-specific constraints for variables $\boldsymbol{x}$. The set of all feasible solutions is denoted as:
\begin{equation}
\label{eq:ss}
\begin{split}
    S= \{\boldsymbol{s}=\{(x_1,d_1),...,(x_n,d_n)\}| \\
        \quad d_i \in D_i, c(\boldsymbol{x}, P) \leq 0\}
\end{split}
\end{equation}
where we omit the subscript of $c(\boldsymbol{x}, P)$ for clarity. The optimal solution of a COP instance $\boldsymbol{s}^* \in S$ should satisfy $f(\boldsymbol{s}^*, P) \leq f(\boldsymbol{s}, P), \scalebox{0.5}{\quad}\forall \boldsymbol{s} \in S$. Typical COPs such as TSP, VRP, KP are well-known NP-hard problems due to their inherent computational complexity, making them difficult to solve optimally. The COPs involved in the paper are further detailed in Appendix A.

\subsection{Transformer for COPs}

Transformer can be used to construct solutions to COPs \cite{kool2018attention,kwon2020pomo,zhou2023towards}.
Concretely, the numerical features of a COP instance $\mathcal{G}^P$ are encoded by an embedding layer, and then updated by an encoder. The global representation learned by the encoder, along with the context representation (e.g., the embedding of the partial tour in construction), is taken as input to the decoder. Next, the decoder is used to iteratively output the probabilities of candidate nodes. The decoding procedure ends when all nodes are selected one by one, according to the probabilities at each iteration. The probabilistic chain rule for constructing a solution $\pi$ is $p_{\boldsymbol{\theta}}(\pi|\mathcal{G}^P)=\prod_{t=1}^{t_{f}} {p_{\theta}(\pi_t|\mathcal{G}^P, \pi_{<t})}$, where $t_f$ refers to the number of total iterations; $\pi_t$ and $\pi_{<t}$ denote the selected node and the current partial solution at the iteration $t$. The Transformer is often trained by REINFORCE algorithm \cite{williams1992simple} with the gradients computed by:
\begin{multline}
\label{eq:reinforce}
    \nabla_{\boldsymbol{\theta}} J(\boldsymbol{\theta}|\mathcal{G}^P) \approx 
    \mathbb{E}_{p_{(\boldsymbol{\theta})}(\pi \mid \mathcal{G}^P)}[(f(\pi)-b(\mathcal{G}^P))\\
    \nabla_{\boldsymbol{\theta}}\log p_{(\boldsymbol{\theta})}(\pi \mid \mathcal{G}^P)]
\end{multline}
where $\mathbb{E}_{p_{(\boldsymbol{\theta})}(\pi \mid \mathcal{G}^P)}f(\pi)$ is the expected cost of solutions and $b(\cdot)$ is a baseline for reducing estimation variance. We refer interested readers to Appendix B for more details on Transformer for COPs.

\begin{figure*}[htb]
\centering
  \includegraphics[width=\textwidth]{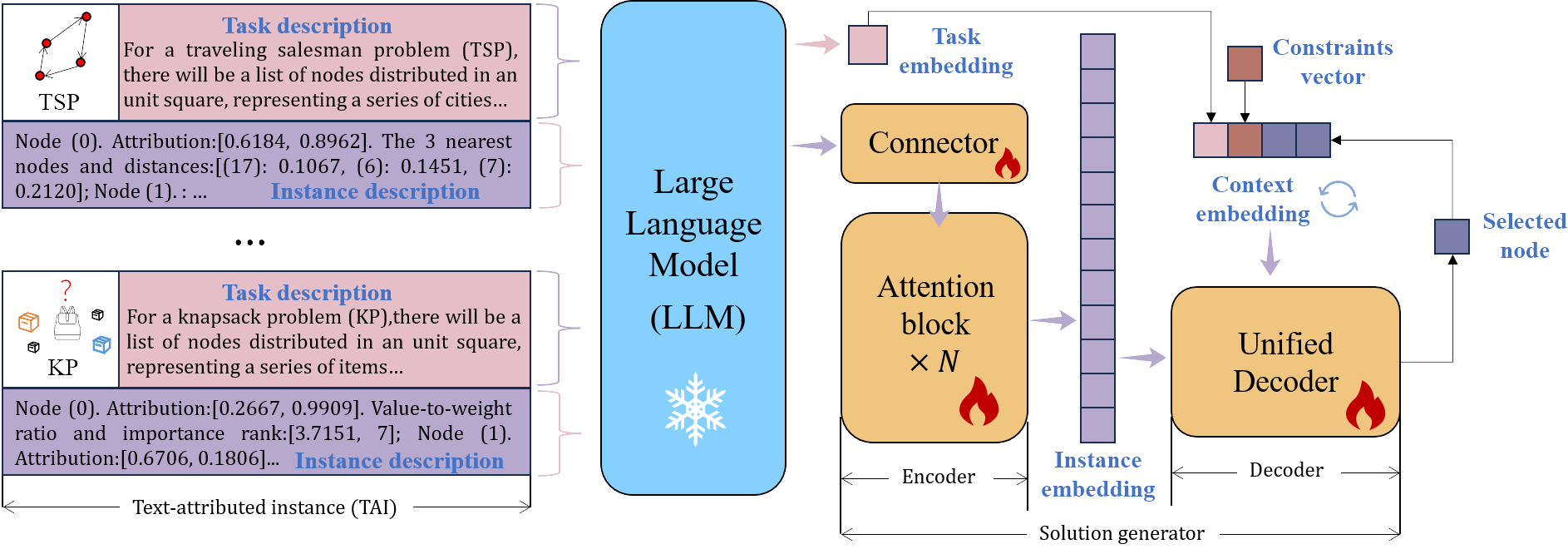}
\caption{The illustration of the proposed framework. [\textit{Blue part}]: The LLM is frozen and takes as input the TAI for different COPs, producing task embedding and initial node embedding. [\textit{Orange part}]: The encoder of the trainable solution generator processes the embedding through the attention blocks and produces instance embeddings, which is further used to construct solutions by a decoder.}
\label{fig:framework}
\end{figure*}

\section{Methodology}

\subsection{TAIs of COPs}
Despite that the formations of various COPs (as displayed in Eq.~(1)) would be different, they can generally be described in natural language.
As shown in Figure~\ref{fig:framework}, we propose to represent COP instances as text-attributed instances (TAIs) through task and instance descriptions. 
The task description specifies the formation of a COP, including decision variables, general constraints, and the objective function, while the instance description specifies the details of nodes or edges on the COP graph.

Formally, the TAI of a COP $P$ is denoted as $\mathcal{I}(\mathcal{G}^P)=\{\kappa^P, v^P\}$, which encapsulates both the task description $\kappa^P$ and the instance description $v^P$, thus enabling the recognition of different COP instances. The task description $\kappa^P$ aims to prompt the LLM with the definition and background of the target COP, while the instance description $v^P$ offers details in an instance, e.g., features on nodes.

Taking the instance description in Figure~\ref{fig:framework} as the example, the sentences $v_i^P \in v^P, i\in \{1, \ldots, n\}$ delineates the numerical attributes on the nodes of an instance, indicating the basic features such as the node coordinates in TSP or the weight-profit pairs in KP. Additionally, we also incorporate heuristic information. By doing so, we aim to prompt the LLM to better understand each instance from the perspective of general and conventional heuristics. For example, according to greedy heuristics, we delineate the top-$k$ ($k=3$) nearest nodes of each node in the instance description of TSP, and the value-to-weight ratios with their ranks in KP instances. The inclusion of heuristic information enhances the LLMs' ability to comprehend the instances, which will be shown in the ablation study.

\subsection{Language-based Neural COP Solver}
Since LLMs cannot effectively solve COPs relying solely on their internal knowledge, the proposed LNCS integrates a Transformer with an LLM, as illustrated in Figure~\ref{fig:framework}. The task and instance descriptions are individually taken as input by the LLM, with their embeddings obtained from the last layer of the LLM. 
Specifically, both the instance and the task description are processed by the LLM, which is expressed by $x_i^P=\text{LLM}(v_i^P)$ and $k^P=\text{LLM}(\kappa^P)$. The embeddings $\{x_i^P\}_{i=1}^n$ contain the information pertaining to the instance, while task embedding $k^P$  reflects the domain-specific information of the COP $P$. 


Inspired by AM \cite{kool2018attention}, we develop a Transformer network in LNCS, which is shown in Figure~\ref{fig:framework}, connecting to an LLM and serving as a solution generator for various text-attributed COPs. As the embeddings of different COPs are aligned within the same semantic space by the LLM, a single Transformer can easily process these embeddings without requiring any additional problem-specific modules. We elaborate the components of the solution generator as follows:

\noindent\textbf{Connector.} Given the embeddings of TAIs, we use a linear projection layer to connect the LLM and the following attention blocks. As the embeddings produced by an LLM are typically high-dimensional, the connector with parameters $\mathbf{W}^e \in \mathbb{R}^{d_o \times d_h}$ and $\mathbf{b}^e \in \mathbb{R}^{d_h}$ is used for dimensionality reduction. As such, the instance embeddings $\{x_i^P\}_{i=1}^n$ are concatenated into $\mathbf{x}^P\in \mathbb{R}^{n \times d_o}$ and then linearly transformed by $\mathbf{h}^{(0)}=\mathbf{W}^e \mathbf{x}^P+\mathbf{b}^e$ with $\mathbf{h}^{(0)} \in \mathbb{R}^{n \times d_h}$.

\noindent\textbf{Encoder.} $\mathbf{h}^{(0)}$ is processed by successive attention blocks, each of which consists of a multi-head attention (\textbf{MHA}) layer~\cite{vaswani2017attention}, a node-wise FeedForward (\textbf{FF}) layer, a skip-connection layer ~\cite{He_2016_CVPR} and a batch normalization (\textbf{BN}) layer~\cite{pmlr-ioffe15}, that is,
\begin{equation}
    \hat{\mathbf{h}}^{(l)} = \textbf{BN}_{l}(\mathbf{h}^{(l-1)}+\textbf{MHA}_{l}(\{\mathbf{h}^{(l-1)}_1,...,\mathbf{h}^{(l-1)}_{n}))
\end{equation}
\begin{equation}
    \mathbf{h}^{(l)} = \textbf{BN}_{l}(\hat{\mathbf{h}}^{(l)}+\textbf{FF}_l(\hat{\mathbf{h}}^{(l)}))
\end{equation}
where $l \in \{1, \dots, N\}$ represents the index of the attention block. The details of $\textbf{MHA}$, $\textbf{FF}$, and $\textbf{BN}$ are elaborated in Appendix C. 
After $N$ attention blocks, the instance embeddings of a TAI are advanced to $\mathbf{h}^{(N)}\in \mathbb{R}^{n \times d_h}$.

\noindent\textbf{Decoder.} 
The decoding is sequentially unfolded. In each step $t \in \{1,...,n\}$, a node is chosen to be appended to the partial solution until a feasible solution is constructed. The decoder should construct the solution for diverse COPs by 1) discriminating different COPs and 2) using a unified decoding process for various COPs. To this end, we need to define the decoding context to incorporate problem-specific information, which is then used to guide specific decoding policies for different COPs.

One feasible way for the decoder to distinguish between different COPs is to provide it with the task description and problem-specific constraints. Given that some COPs entail dynamic constraints, such as the constraints defined by the varying vehicle load and knapsack capacity in the Capacitated VRP (CVRP) and KP, we first incorporate these dynamic attributes into the decoding context. We use a variable $c^P_t$ to monitor these dynamic attributes, indicating the remaining vehicle load $C^v_t$ or knapsack capacity $C^b_t$ at each decoding step $t$, such that:
\begin{equation}
    c^P_t = 
    \begin{cases}
    C^v_t, \quad \text{if the COP is CVRP,} \\
    C^b_t, \quad \text{if the COP is KP,} \\
    0, \quad \scalebox{0.5}{\quad} \text{otherwise}
    \end{cases}
\end{equation}
Note that we only take CVRP and KP as examples. One can easily extend it to encompass dynamic attributes of constraints in other COPs.

Besides the dynamic attribute $c^P_t$ and task embedding $k^P$, which involve task-specific information in the decoding context, the probability of selecting a node $\pi_t$ at step $t$ should also be associated with the state of the current partial solution. Therefore, the final decoding context constitutes $\mathbf{c}^P_t$, $k^P$, $\pi_{<t}$. Following \cite{kool2018attention}, we use the embeddings of the first and last selected node, i.e., $\mathbf{h}^{(N)}_1$ and $\mathbf{h}^{(N)}_t$, to represent the partial solution $\pi_{<t}$. The context is formulated as:
\begin{equation}
    \mathbf{h}_{(c)}^t = 
    \begin{cases}
    [\mathbf{c}^P_t, k^P, \mathbf{h}^{(N)}_1, \mathbf{h}^{(N)}_t], \text{if }t>1 \text{,} \\
    \text{none}, \qquad\qquad\qquad \scalebox{0.4}{\quad} \text{if }t=1
    \end{cases}
\end{equation}
where $[\cdot,\cdot,\cdot,\cdot]$ indicates a horizontal concatenation operation. Specially, we do not use the context at the first decoding step. Instead, we specify all nodes to start constructing $n$ solutions in parallel, which facilitate the exploration of solution space for performance enhancement~\cite{kwon2020pomo}.

Subsequently, the context $\mathbf{h}^t_{(c)}$ is taken as the query in an \textbf{MHA} layer, while the key and value are derived from $\mathbf{h}^{(N)}$ by linear transformations. The compatibility between the decoding context $\mathbf{h}^t_{(c)}$ and the key is calculated by:
\begin{equation}
   \hat{\mathbf{h}}_{(c)}^t=\textbf{MHA}(\{\mathbf{h}^t_{(c)}, \mathbf{h}^{(N)}, \mathbf{h}^{(N)}\}) 
\end{equation}

The logit for selecting node $i$ is calculated by:
\begin{equation}
    u_i = C \cdot \tanh(\frac{\mathbf{\hat{\mathbf{h}}_{(c)}^t}\mathbf{h}_{i}}{\sqrt{d_A}}), i=1,2,\ldots,n
\end{equation}
where $d_A$ is the hidden dimension of the \textbf{MHA} layer; $C=10$ is the parameter for logit clipping; $\mathbf{h}_{j}$ is the embedding of node 
$i$ in $\mathbf{h}^{(N)}$. Provided that the constraints cannot be satisfied by selecting node $i$ (e.g., the constraint is violated by $i = \pi_{t^{'}},\scalebox{0.4}{\quad} \forall t^{'} < t$ for TSP), the corresponding logit $u_i$ is set to $-\infty$. Finally, the decoder selects a node $\pi_t$ at step $t$ by sampling from according to $p(\pi_t)=\text{Softmax}(\{u_i\}_{i=1}^{n})$. The decoder iteratively selects nodes to construct a feasible solution until no more candidate nodes remain. 


\subsection{Training Scheme}
\begin{figure}[t]
\centering
  \includegraphics[width=0.38\textwidth]{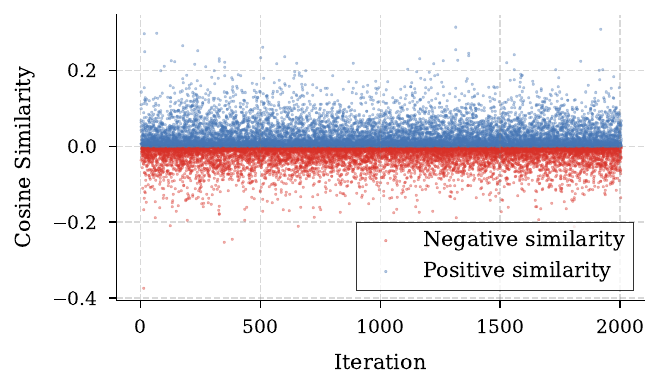}
\caption{The cosine similarities between gradients of loss functions for 5 COPs (with $n=50$). The gradients are calculated during training the LNCS by 2000 batches, following the vanilla averaged REINFORCE.}
\label{fig:grads}
\end{figure}

We resort to RL to train the solution generator in the LNCS, while freezing the parameters of the LLM. Treating each COP as a task, one approach to training with multiple tasks is to average their RL losses for gradient backpropagation \cite{liu2024multi}. 
However, this simple averaging scheme for clearly different types of COPs presents a challenge in optimizing neural networks \cite{liu2021conflict}. It may yield undesirable outcome for two reasons: Firstly, the objectives of different COPs may exhibit varying scales, making the larger gradients dominate the update of the model. Secondly, the updated directions of parameters may conflict under different objectives, potentially leading to performance degradation for specific COPs. These limitations will compromise the ability of LLMs to unify semantic representation for different COPs. To showcase the issue, we calculate the cosine similarities to determine if the gradients between tasks are contradictory, with the results plotted in Figure~\ref{fig:grads}. As observed, the update directions of gradients for different COPs indeed have considerable conflicts, i.e., negative cosine similarity, which potentially harms the training of the model.

To overcome the issue, we use a simple yet effective multi-task RL algorithm, named conflict gradients erasing reinforcement learning (CGERL), to train the LNCS. Inspired by \cite{yu2020gradient}, we project the conflicting gradients onto the normal plane of others, thereby enabling the model to share common knowledge from different COPs in a favorable conflict-free manner. Formally, suppose that we have a set of tasks $\{P_i\}_{i=1}^{\mathcal{N}}$ denoting $\mathcal{N}$ different COPs. Let $\mathbf{g}_i$ and $\mathbf{g}_j$ denote the gradients of tasks $P_i \in \{P_i\}_{i=1}^{\mathcal{N}}$ and $P_j \in \{P_i\}_{i=1}^{\mathcal{N}}$, respectively. Each of the gradients can be estimated by Eq.~(\ref{eq:reinforce}), following the typical REINFORCE algorithm \cite{williams1992simple}. Afterward, we compute $\delta = \mathbf{g}_i \cdot \mathbf{g}_j$ to determine whether the two gradients are conflicting ($\delta<0$) or not ($\delta \geq 0$). As such, the gradient for task $P_i$ can be adapted by:
\begin{equation}
\label{eq:gradient}
    \hat{\mathbf{g}}_i = 
    \begin{cases}
        \mathbf{g}_i - \frac{\mathbf{g}_i \cdot \mathbf{g}_j}{\Vert \mathbf{g}_j \Vert^2} \mathbf{g}_j  \text{, \scalebox{0.2}{\quad} if }\delta<0 \text{,} \\
        \mathbf{g}_i \text{, \qquad \qquad \quad if }\delta \geq 0
    \end{cases}
\end{equation}
The conflict elimination process is conducted for each training batch, consisting of multiple tasks. For each target task $P_i \in \{P_i\}_{i=1}^{\mathcal{N}}$, we sample other tasks $P_j$, $\forall j \neq i$ in random order and progressively apply Eq.~(\ref{eq:gradient}). The process repeats until the gradients of all tasks are adapted. Finally, the adapted gradients are aggregated as $\mathbf{g}=\sum_i^\mathcal{N} \hat{\mathbf{g}}_i$, which is used to update the solution generator. The overall training procedure is provided in Appendix D.

\section{Experiments}
We mainly use Llama2-7b \cite{touvron2023llama} as the LLM for the evaluation of the method. We also provide an ablation study to test the performance of other language models. More details of the experiment settings, including model configuration and training process are provided in Appendix E.

\subsection{Benchmarks}
The proposed LNCS is evaluated on five representative COPs, including TSP, CVRP, KP, minimum vertex cover problem (MVCP), and single-machine total weighted tardiness problem (SMTWTP). Meanwhile, we fine-tune the trained LNCS on two new tasks, including the VRP with backhauls (VRPB) and the maximum independent set problem (MISP). The instance generation for the COPs and the examples of their TAIs are provided in Appendix F. Note that many other COPs can also be incorporated besides the above problems.

\subsection{Baselines}
\textbf{LLM-based Optimization:} We first compare our NLCS approach with other LLM-based methods, encompassing both \textit{LLMs as programmers} and \textit{LLMs as optimizers}, as outlined below:

We use Algorithm Evolution Using LLMs (AEL) \cite{liu2023algorithm}, Reflective Evolution (ReEvo) \cite{ye2024reevo}, and Self-Guiding Exploration (SGE) \cite{iklassov2024self} as baselines for \textit{LLMs as programmers}, which leverage embedded knowledge to generate heuristics to solve COP. AEL and ReEvo are utilized to evolve constructive heuristics for solving the TSP, while ReEvo is also used to improve the ant colony optimization (ACO) for solving CVRP and KP. In addition, we include LLM-driven Evolutionary Algorithms (LMEA) \cite{liu2023large11} and Optimization by PROmpting (OPRO) \cite{yang2023large} as baselines for \textit{LLMs as optimizers}, which attempt to directly produce solutions from textual problem descriptions.

\noindent\textbf{Traditional Solvers:} We employ OR-Tools, which integrates various heuristic approaches, to solve TSP, CVRP, and KP. Additionally, we also use the Gurobi solver to obtain optimal solutions for the COPs and accordingly calculate the optimality gaps to evaluate other methods. 

The LNCS is also compared with conventional heuristics, such as the nearest neighbor and the farthest insertion for TSP, the sweep heuristic and parallel saving algorithm for CVRP \cite{rasku2019meta}, the greedy policy for KP, the MVCApprox method (i.e., greedily reduces costs over edges and iteratively builds a cover) \cite{bar1985local} and the randomized edge-based heuristic (REH) \cite{pitt1985simple} for MVCP and the earliest due date (EDD) dispatching rule \cite{jackson1955scheduling} for SMTWTP. We also use the ACO (with 20 ants and 50 iterations) \cite{ye2024deepaco} as the metaheuristic baseline to solve the problems. The heuristics for TSP and CVRP are based on the implementation of pyCombinatorial\footnote{https://github.com/Valdecy/pyCombinatorial} and VeRyPy\footnote{https://github.com/yorak/VeRyPy}, respectively.

\noindent\textbf{NCO Solvers:} We compare LNCS with other NCO approaches, including AM \cite{kool2018attention} and POMO \cite{kwon2020pomo}, across TSP, CVRP, and KP. These methods rely on numerical features as input, enabling us to assess the current performance gap between text-attributed COP solutions and numerically based NCO methods.

\subsection{Performance Evaluation}

\noindent\textbf{Comparison with LLM-based Optimization.} To evaluate our method, we compare it against both \textit{LLMs as programmers} and \textit{LLMs as optimizers} approaches, which have been extensively explored in the LLM community. We center our comparison on TSP, CVRP, and KP. The results presented in Table~\ref{tab:comparison1} demonstrate that LNCS achieves the best performance among all LLM-based frameworks. 


Although existing methods can only address small-scale COPs effectively \cite{yang2023large}, their performance still remains suboptimal even on the simple TSP (with 20 nodes), suggesting that relying solely on embedded knowledge within LLMs is insufficient at the current stage. In contrast, our LNCS incorporates an additional Transformer and a specialized RL training process, allowing the integration of domain-specific knowledge of COPs. This enhancement enables LLMs to solve various COPs more effectively in a unified way.

\begin{table}
\renewcommand{\arraystretch}{0.7}
  \centering
{\fontsize{10}{11}\selectfont
  \begin{tabular}{llccc}
    \hline
    \multirow{6}{*}[-3ex]{\rotatebox{90}{\scriptsize{\emph{TSP}}}}  & Method & $n=20$ & $n=50$ & $n=100$ \\
    \hline
    & AEL     & 7.78\%     &    10.50\%     & 12.35\%  \\
    & ReEvo     & 7.77\%    &   10.23\%     & 11.87\%    \\
    & SGE     & 11.32\%    &   45.28\%      &  -    \\
    & $\text{LMEA}^*$     & 3.94\%     &    -   &   -      \\
    & $\text{ORPO}^*$       & 4.40\%     &   133.0\%    &   -       \\
    & LNCS (ours)     & \textbf{0.39\%}       &  \textbf{1.64\%}   &  \textbf{4.38\%} \\
    \hline
    \multirow{3}{*}[-0.6ex]{\rotatebox{90}{\scriptsize{\emph{CVRP}}}} &
    ReEvo     & 5.19\%    &   14.27\%     & 19.59\%    \\
    & SGE     & 76.46\%    &   144.21\%      &  -    \\ 
    & LNCS (ours)     & \textbf{2.51\%}       &  \textbf{3.62\%}   &  \textbf{5.59\%} \\
    \hline
    \multirow{3}{*}[-0.6ex]{\rotatebox{90}{\scriptsize{\emph{KP}}}} &
    ReEvo     & 0.14\%    &   4.31\%     & 9.40\%    \\
    & SGE     & 42.62\%    &   39.08\%      &  -    \\ 
    & LNCS (ours)     & \textbf{0.10\%}       &  \textbf{0.06\%}   &  \textbf{0.03\%} \\
    \hline
  \end{tabular}
  }
  \caption{The optimality gaps of LLM-based approaches on TSP. *: Results are drawn from the original literature and based on  gpt-3.5-turbo. -: Excessively long computation time leads to the unavailability of results.}
  \label{tab:comparison1}
\end{table}

\noindent\textbf{Comparison with Heuristic-Based Optimization.} Given that the LNCS achieves state-of-the-art (SOTA) performance among existing LLM-based methods, we further investigate its performance relative to traditional heuristic methods. The results of the comprehensive experiments are presented in Table~\ref{tab:routing}, where the LNCS generally outperforms the heuristic baselines, particularly for instances where $n \in \{50, 100\}$. These findings indicate that our approach, through additional training, effectively addresses the LLM's limitations in solving larger COPs (i.e. instances with over 50 nodes).

\begin{table*}[t!]
    \centering
  {\fontsize{9}{11}\selectfont
    \begin{tabular}{ll|ccc|ccc|ccc}
        \hline
        \multirow{5}{*}[-6ex]{\rotatebox{90}{\scriptsize{\emph{TSP}}}} & \multirow{2}{*}{Method} & \multicolumn{3}{c}{$n=20$} & \multicolumn{3}{c}{$n=50$} & \multicolumn{3}{c}{$n=100$} \\
        & & Obj. & Gap & Time & Obj. & Gap & Time & Obj. & Gap & Time \\
        \cline{0-10}
         & OR tools  & 3.85 & 0.00\% & 0.36s  & 5.87 & 3.07\% & 0.60s & 8.13 & 4.65\% & 1.32s \\
         & Nearest neighbor  & 4.51 & 17.21\% & $<$0.01s & 6.98 & 22.72\% & 0.03s & 9.69 & 24.86\% & 0.10s \\
         & Farthest insertion  & 3.96 & 2.89\% & 0.21s & 5.98 & 4.97\% & 4.32 & 8.21 & 5.74\% & 126s \\
         & ACO & 3.94 & 2.23\% & 0.74s & 6.54 & 14.94\% & 1.53s & 9.99 & 28.74\% & 2.01s \\
         & LNCS (ours) & 3.87 & 0.39\% & 0.72s  & 5.79 & 1.64\% & 1.64s & 8.10 & 4.38\% & 3.60s \\
        \hline
        \multirow{5}{*}[-1ex]{\rotatebox{90}{\scriptsize{\emph{CVRP}}}} &
         OR tools  & 6.18 & 1.30\% & 0.27s  & 11.05 & 6.63\% & 0.48s & 17.36 & 12.07\% & 1.40s \\
         & Sweep heuristic   & 7.51 & 23.17\% & 0.01s  & 15.65 & 50.95\% & 0.05s & 28.40 & 83.39\% & 0.25s \\
         & Parallel saving  & 6.33 & 3.85\% & $<$0.01s  & 10.90 & 5.18\% & $<$0.01s & 16.42 & 6.03\%  & 0.03s \\
         & ACO  & 7.72 & 26.51\% & 0.80s  & 15.76 & 52.12\% & 1.97s & 26.66 & 72.07\%  & 4.90s \\
         & LNCS (ours) & 6.25 & 2.51\% & 0.90s  & 10.74 & 3.62\% & 2.15s & 16.35 & 5.59\% & 4.80s \\
        \hline
         \multirow{4}{*}[-1ex]{\rotatebox{90}{\scriptsize{\emph{KP}}}} &
         OR tools  & 7.948 & -0.01\% & $<$0.01s & 20.086 & -0.01\% & $<$0.01s & 40.377 & 0.00\% & $<$0.01s \\
         & Greedy policy  & 7.894 & 0.67\% & $<$0.01s  & 20.033 & 0.26\% & $<$0.01s & 40.328 & 0.12\% & $<$0.01s \\
         & ACO & 7.947 & 0.00\% & 0.72s  & 20.053 & 0.15\% & 2.19s & 40.124 & 0.62\% & 3.41s \\
         & LNCS (ours) & 7.939 & 0.10\% & 0.06s  & 20.071 & 0.06\% & 0.17s & 40.361 & 0.03\% & 0.26s \\
        \hline
        \multirow{3}{*}[-0.5ex]{\rotatebox{90}{\scriptsize{\emph{MVCP}}}} &
         MVCApprox   & 14.595 & 22.13\% & $<$0.01s  & 34.856 & 20.98\% & $<$0.01s & 68.313 & 21.57\% & $<$0.01s \\
         & REH   & 16.876 & 41.22\% & $<$0.01s  & 41.426 & 43.78\% & $<$0.01s & 81.860 & 45.68\% & $<$0.01s \\
         & LNCS (ours) & 12.900 & 7.93\% & 0.1s  & 32.101 & 11.42\% & 0.43s & 64.893 & 15.49\% & 1.63s \\
        \hline
        \multirow{3}{*}[-0.5ex]{\rotatebox{90}{\scriptsize{\emph{SMTWTP}}}} &
         EDD  & 0.3822 & 275.81\% & $<$0.01s  & 0.4461 & 107.68\% & $<$0.01s & 0.4434 & 81.87\% & $<$0.01s \\
         & ACO  & 0.2967 & 191.74\% & 0.35s  & 1.0471 & 387.48\% & 1.35s & 6.77 & 2677\% & 2.00s \\
         & LNCS (ours) & 0.2862 & 181.41\% & 0.09s  & 0.3353 & 56.10\% & 0.31s & 0.3316 & 36.01\% & 1.10s \\
        \hline
        
    \end{tabular}
   }
    \caption{Performance on 1K instances for the COPs. Obj. indicates the average objective values. The gaps are computed using the optimal solutions produced by Gurobi. The average gaps and computation time are reported.}
    \label{tab:routing}
\end{table*}

\noindent\textbf{Comparison with NCO.} We also compare our model with typical NCO models. The results can be found in Appendix G, where we find that LNCS still achieves satisfactory performance.

\begin{figure}[htb]
\centering
    \subfloat[]{\includegraphics[width=0.22\textwidth]{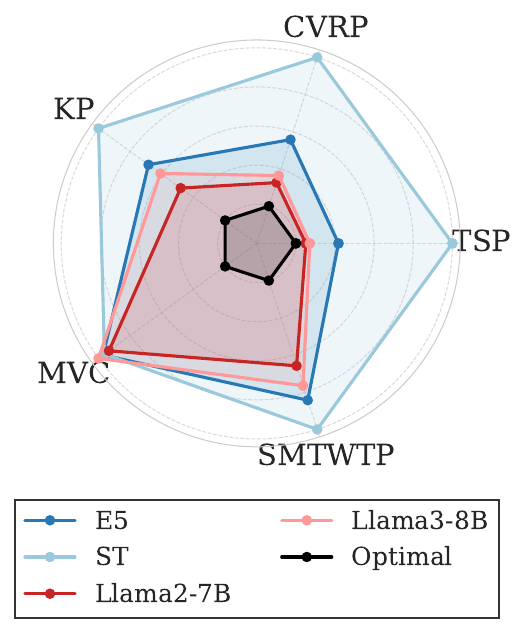}}\hfill
    \subfloat[]{\includegraphics[width=0.22\textwidth]{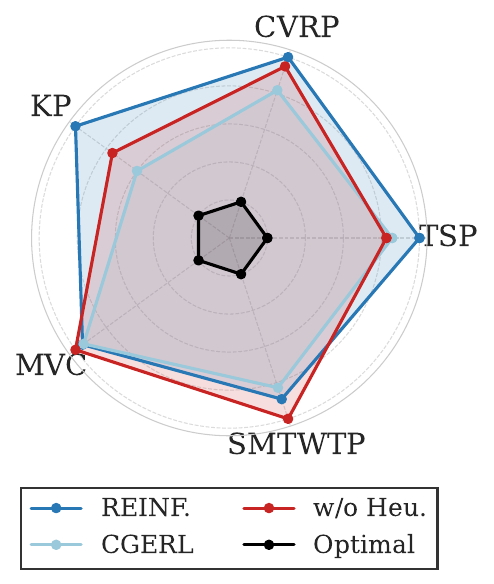}}
\caption{Performance comparison (with $n=50$) of the LNCS under different settings: (a) different LLMs (b) different training algorithms. The smaller the shadow area is, the better the corresponding setting performs.}
\label{fig:abalation}
\end{figure}

\subsection{Ablation Study}
\label{sec:ablation}

\noindent\textbf{Different LLMs. }We utilize different language models for encoding TAIs, including sentence transformer (ST) \cite{reimers2019sentence}, e5-large-v2 (E5) \cite{wang2022text}, Llama2-7b, and Llama3-8b. The normalized results are compared in Figure~\ref{fig:abalation} (a). We find that performance can generally be improved with an increase in the scale of language models. An exception is that Llama2-7b outperforms Llama3-8b in all tasks, indicating that advanced LLMs do not necessarily lead to better learning outcome.

\noindent\textbf{Effect of CGERL. }We train the LNCS by both CGERL and vanilla REINFORCE (REINF.), and the normalized results are compared in Figure~\ref{fig:abalation} (b). LLM with RL that has undergone conflict gradient elimination generates better solutions than the original one, demonstrating the necessity for introducing effective multi-task learning methods.

\noindent\textbf{Heuristic Information in TAI. }We also train the model with and without heuristic information in TAI, and the results are also presented in Figure~\ref{fig:abalation} (b). We find that the additional information added to the prompt can further enhance model performance, particularly on tasks such as CVRP, KP, and SMTWTP, indicating that prompt engineering can also slightly improve the LNCS.

\noindent\textbf{Synergistic Learning between Tasks.} We empirically find that the LLM has facilitated synergistic learning between different COPs, and the results are shown and analyzed in Appendix G.

\subsection{Model Generalizability}

Generalizability is a key advantage offered by LLMs that enables rapid adaptation to downstream tasks. To assess this capability, we evaluate the generalizability by fine-tuning LNCS on new tasks.

\noindent\textbf{Fine-tuning on New COPs. }We evaluate the cross-problem transfer capability of the trained LNCS by fine-tuning it on new COPs, i.e., VRPB and MISP. We use the model trained on COPs with $n=50$ and fine-tune it on VRPB with $n=50$ (VRPB50) and MISP with $n=100$ (MISP100), respectively.

\begin{figure}[thb]
    \centering
    \subfloat[]{\includegraphics[width=0.237\textwidth]{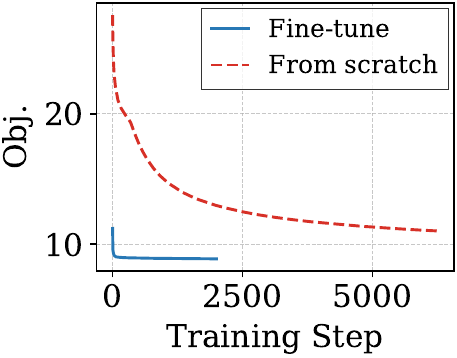}}\hfill
    \subfloat[]{\includegraphics[width=0.223\textwidth]{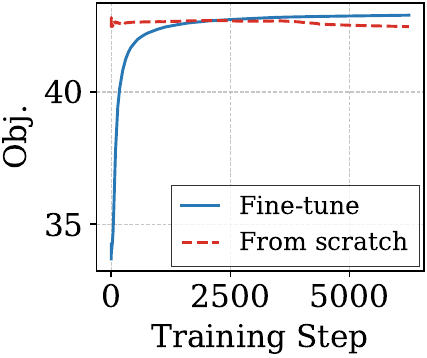}}\hfill
    \caption{Results by fine-tuning and learning from scratch for (a) VRPB50 and (b) MISP100.}
    \label{fig:new_task}
\end{figure}

We randomly initialize the LNCS and train it on each task using 6,000 steps from scratch, which is taken as the baseline. In contrast, for the VRPB50 task, we employ only 2,000 steps for fine-tuning, while for the MISP100 task, the fine-tuning process follows the same number of steps as in the training from scratch. The results are shown in Figure~\ref{fig:new_task}. As observed, for the VRPB task, the trained LNCS already learned to solve some routing problems, such as CVRP and TSP, so that it can get better result than random initialization even in the first batch. Compared to learning from scratch, the fine-tuning on new VRP tasks brings about a faster convergence with better performance, demonstrating the few-shot generalization capability of the model.

For the MISP task, the convergence is slower than that of learning from scratch. This is primarily because MISP and MVCP are complementary problems on any given graph. Specifically, for a graph $G=(V, E)$, the maximum independent set is $I = V/C$, where $C$ is the maximum vertex cover of $G$. Consequently, a neural network trained on MVCP must learn an entirely different decision-making process for MISP, leading to suboptimal performance in the initial batches. However, after approximately 2,000 steps, the fine-tuning achieves performance comparable to that of the learning-from-scratch model and eventually converges to a better result. This shows that fine-tuning on a complementary task can still be beneficial.

\noindent\textbf{Fine-tuning on New Size. }We compare the fine-tuning and training from scratch for COPs with $n=100$. The results shown in Appendix G demonstrate that fine-tuning on new problem size allows rapid adaptations under few-shot setting.

\section{Conclusion}
This paper presents a framework for elevating LLMs in solving text-attributed COPs. Leveraging TAIs, the method integrates an LLM with a Transformer-based solution generator. To enable simultaneous learning across multiple tasks, a multi-task RL approach is employed to address conflicting gradients during training. The proposed model demonstrates competitive performance across various COPs, surpassing all existing LLM-based optimization methods. Consequently, this work introduces a unified paradigm to enable LLMs to generate high-quality COP solutions.

\ifCLASSOPTIONcaptionsoff
  \newpage
\fi



%

\bibliographystyle{IEEEtran}
\bibliography{IEEEexample, IEEEabrv}

\end{document}